\pdfoutput=1
\documentclass{article}

\usepackage{arxiv}

\usepackage[utf8]{inputenc} 
\usepackage[T1]{fontenc}    
\usepackage{hyperref}       
\usepackage{url}            
\usepackage{booktabs}       
\usepackage{amsfonts}       
\usepackage{nicefrac}       
\usepackage{microtype}      
\usepackage{graphicx}
\usepackage{doi}
\usepackage{xcolor}
\usepackage{caption}
\usepackage{subcaption}
\usepackage{amsmath}
\usepackage{float}
\usepackage[numbers]{natbib}

\newcommand{\dd}[2]{\frac{\partial{#1}}{\partial{#2}}}

\title{A Robust Initialization of Residual Blocks for Effective ResNet Training without Batch Normalization}

\author{ 
    Enrico Civitelli, Alessio Sortino \& Matteo Lapucci \\
	Dipartimento di Ingegneria dell’Informazione \\
	Università degli Studi di Firenze \\
	Via di S. Marta 3, 50139, Firenze, Italy \\
	\texttt{\{enrico.civitelli,alessio.sortino,matteo.lapucci\}@unifi.it} \\
	\And
	Francesco Bagattini \& Giulio Galvan \\
	Flair Tech \\
	Firenze, Italy \\
	\texttt{\{francesco.bagattini,giulio.galvan\}@flair-tech.com} \\
}

\date{}

\renewcommand{\shorttitle}{NormalizationFree ResNet}

\hypersetup{
    pdftitle={A Robust Initialization of Residual Blocks for Effective ResNet Training without Batch Normalization},
    pdfsubject={NormalizationFree Residual Network},
    pdfauthor={Enrico Civitelli, Alessio Sortino, Matteo Lapucci, Francesco Bagattini and Giulio Galvan},
    pdfkeywords={Normalization-Free ResNets, Weights Initialization, Exploding Gradient, Residual Blocks},
}

\begin{document}
\maketitle

\begin{abstract}
	Batch Normalization is an essential component of all state-of-the-art neural networks architectures. However, since it introduces many practical issues, much recent research has been devoted to designing normalization-free architectures. In this paper, we show that weights initialization is key to train ResNet-like normalization-free networks. In particular, we propose a slight modification to the summation operation of a block output to the skip-connection branch, so that the whole network is correctly initialized. We show that this modified architecture achieves competitive results on CIFAR-10, CIFAR-100 and ImageNet without further regularization nor algorithmic modifications.
\end{abstract}

\section{Introduction} \label{sec:introduction}
Batch normalization \citep{ioffe2015batch}, in conjunction with skip connections \citep{he2016deep, he2016identity}, has allowed the training of significantly deeper networks, so that most state-of-the-art architectures are based on these two paradigms.

The main reason why this combination works well is that it yields well behaved gradients (removing \textit{mean-shift}, avoiding \textit{vanishing} or \textit{exploding} gradients). As a consequence, the training problem can be ``easily'' solved by SGD or other first-order stochastic optimization methods. Furthermore, batch normalization can have a regularizing effect \citep{hoffer2017, luo2019understanding}.

However, while skip connections can be easily implemented and integrated in any network architecture without major drawbacks, batch normalization poses a few practical challenges. As already observed and discussed by \cite{brock2021characterizing, brock2021highperformance} and references therein, batch normalization adds a significant memory overhead, introduces a discrepancy between training and inference time, has a tricky implementation in distributed training, performs poorly with small batch sizes \citep{yan2020towards} and breaks the independence between training examples in a minibatch, which can be extremely harmful for some learning tasks \citep{lee2020residual, lomonaco2020rehearsal}. 

For these reasons a new stream of research emerged which aims at removing batch normalization from modern architectures. Several works \citep{zhang2019, soham2020batch, bachlechner2020rezero} aim at removing normalization layers by introducing a learnable scalar at the end of the residual branch, i.e., computing a residual block of the form $x_{l} = x_{l-1} + \alpha f(x_{l-1})$. The scalar $\alpha$ is often initialized to zero so that the gradient is dominated, early on in the training, by the skip path. While these approaches have been shown to allow the training of very deep networks, they still struggle to obtain state-of-the-art test results on challenging benchmarks. 

\cite{shao2020normalization} propose a different modification of the standard residual layer, suitably carrying out a weighted sum of the identity and the non-linear branches. 

More recently \cite{brock2021characterizing, brock2021highperformance} proposed an approach that combines a modification of the residual block with a careful initialization, a variation of the Scaled Weight Standardization \citep{hang2017centered, qiao2020microbatch} and a novel adaptive gradient clipping technique. Such combination has been shown to obtain competitive results on challenging benchmarks.

In this work we propose a simple modification of the residual block summation operation that, together with a careful initialization, allows to train deep residual networks without any normalization layer. Such scheme does not require the use of any standardization layer nor algorithmic modification. Our contributions are as follows:

\begin{itemize}
	\item We show that while \textit{NFNets} of \cite{brock2021characterizing, brock2021highperformance} enjoy a perfect forward variance (as already noted by \cite{brock2021characterizing}), it puts the network in a regime of \textit{exploding gradients}. This is shown by looking at the variance of the derivatives of the loss w.r.t.\ to the feature maps at different depths. 
	\item We propose a simple modification of the residual layer and then develop a suitable initialization scheme building on the work of \cite{he2015delving}.
	\item We show that the proposed architecture achieves competitive results on CIFAR-10, CIFAR-100, and ImageNet. \citep{krizhevsky2009learning}, which we consider evidence supporting our theoretical claims.
\end{itemize}

\section{Background} \label{sec:background}
As highlighted in a number of recent studies \citep{hanin2018start, devansh2019initialize, yann2019metainit}, weights initialization is crucial to make deep networks work in absence of batch normalization. In particular, the weights at the beginning of the training process should be set so as to correctly propagate the forward activation and the backward gradients signal in terms of mean and variance. 

This kind of analysis was first proposed by \cite{glorot2010understanding} and later extended by \cite{he2015delving}. These seminal studies considered architectures composed by a sequence of convolutions and Rectified Linear Units (ReLU), which mainly differ from modern ResNet architectures for the absence of skip-connections.

The analysis in \cite{he2015delving} investigates the variance of each response layer $l$ (\textit{forward variance}):
\begin{gather*}
	z_l =\text{ReLU}(x_{l-1}),\qquad
	x_{l} = W_l z_l. 
\end{gather*}
The authors find that if $\mathbb{E}[x_{l-1}]=0$ and $\text{Var}[x_{l-1}]=1$ the output maintains zero mean and unit variance if we initialize the kernel matrix in such a way that:
\begin{equation} \label{eq:he_for}
	\text{Var}[W] = \frac{2}{n_{\text{in}}},
\end{equation}
where $n_{\text{in}}=k^2c$ with $k$ the filter dimension and $c$ the number of input channels (\textit{fan in}).

A similar analysis is carried out considering the gradient of the loss w.r.t.\ each layer response (\textit{backward variance}) $\dd{\cal L}{x_l}$. In this case we can preserve zero mean and constant variance if we have 
\begin{equation} \label{eq:he_back}
	\text{Var}[W] = \frac{2}{n_{\text{out}}},
\end{equation}
where $n_{\text{out}}=k^2d$ with $k$ the filter dimension and $d$ the number of output channels (\textit{fan out}).

Note that equations (\ref{eq:he_for}) and (\ref{eq:he_back}) only differ for a factor which, in most common network architectures, is in fact equal to 1 in the vast majority of layers. Therefore, the initialization proposed by \cite{he2015delving} should generally lead to the conservation of both \textit{forward} and \textit{backward} signals.

The two derivations are reported, for the sake of completeness, in Appendix A.

In a recent work \cite{brock2021characterizing} argued that initial weights should not be considered as random variables, but are rather the realization of a random process. Thus, empirical mean and variance of the weights should do not coincide with the moments of the generating random process. Hence, normalization of the weights matrix should be performed after sampling to obtain the desired moments. Moreover, they argue that channel-wise responses should be analyzed. This leads to the different initialization strategy:
\begin{equation} \label{eq:brock_for}
	\text{Var}[W_i] = \frac{2/(1-\frac{1}{\pi})}{n_{\text{in}}}, 
\end{equation}
where $W_i$ is a single channel of the filter. Note that if mean and variance are preserved channel-wise, then they are also preserved if the whole layer is taken into account.

The authors do not take into account the \textit{backward variance}.
\cite{brock2021characterizing} show that the latter initialization scheme allows to experimentally preserve the channel-wise activation variance, whereas He's technique only works at the full-layer level. 

In the ResNet setting, initialization alone is not sufficient to make the training properly work without batch normalization, if the commonly employed architecture with Identity Shortcuts (see Figure \ref{fig:original_preactivation}) is considered. 

In particular, the skip-branch summation
\begin{equation}
	x_l = x_{l-1} + f_l(x_{l-1}),
\end{equation}
at the end of each block does not preserve variance, causing the phenomenon known as \textit{internal covariate shift} \citep{ioffe2015batch}.

In order to overcome this issue, Batch Normalization has been devised. More recently, effort has been put into designing other architectural and algorithmic modifications that dot not rely on batch statistics. 

Specifically, \cite{zhang2019, soham2020batch, bachlechner2020rezero} modified the skip-identity summation as to downscale the variance at the beginning of training, biasing, in other words, the network towards the identity function, i.e., computing
\begin{equation*}
	x_{l+1}=x_{l-1}+\alpha f_l(x_{l-1}).
\end{equation*}
This has the downside that $\alpha$ must be tuned and is dependent on the number of layers. Moreover, while these solutions enjoy good convergence on the training set, they appear not to be sufficient to make deep ResNets reach state-of-the-art test accuracies \citep{brock2021characterizing}.


Similarly, \cite{shao2020normalization} suggest to compute the output of the residual branch as a weighted sum between the identity and the non-linear branch. Formally, the residual layer becomes
$$x_{l} = \alpha_l x_{l-1} + \beta_l f(x_{l-1}),$$ where coefficients $\alpha_l$ and $\beta_l$ can be set so that the \textit{forward variance} is conserved by imposing that $\alpha_l^2+\beta_l^2=1$. Different strategies can be employed to choose their relative value.

More recently, \cite{brock2021characterizing} proposed to additionally perform a runtime layer-wise normalization of the weights, together with the empirical channel-wise intialization scheme. 
However, we show in the following that the latter scheme, while enjoying perfectly conserved forward variances, induces the network to work in a regime of \textit{exploding gradients}, i.e., the variance of the gradients of the shallowest layers is exponentially larger than that of the deepest ones. 
Reasonably, \cite{brock2021highperformance} found the use of a tailored adaptive gradient clipping to be beneficial because of this reason.

\section{The Proposed Method} \label{sec:proposed_method}
In order to overcome the issue discussed at the end of the previous section, we propose to modify the summation operation of ResNet architectures so that, at the beginning of the training, the mean of either the activations or the gradients is zero and the variance is preserved throughout the network. In our view, our proposal is a natural extension of the work of \cite{he2015delving} for the case of ResNet architectures. Note that, to develop an effective initialization scheme, the residual block summation has to be slightly modified.

\begin{figure}
	\centering
	\begin{subfigure}{0.47\columnwidth}
		\centering
		\includegraphics[height=0.25\textheight]{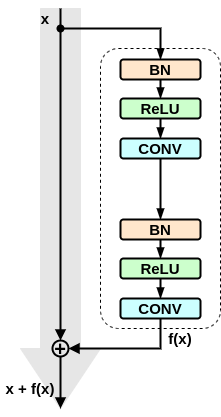}
		\caption{Standard pre-activated Residual Block}
		\label{fig:original_preactivation}
	\end{subfigure}
	\hfill
	\begin{subfigure}{0.47\columnwidth}
		\centering
		\includegraphics[height=0.25\textheight]{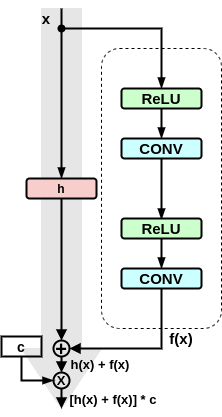}
		\caption{Generalized Normalizer-Free Residual Block}
		\label{fig:proposed_preactivation}
	\end{subfigure}
	\caption{Architectures of Residual Blocks. For both pictures the grey arrow marks the easiest path to propagate the information.}
	\label{fig:original_vs_our_preactivation}
\end{figure}

Namely, we analyze the following general scheme  (see Figure \ref{fig:proposed_preactivation}):
\begin{equation} \label{eq:gen_scheme}
	x_l = c \cdot \left(h(x_{l-1}) + f_l(x_{l-1})\right),
\end{equation}
where $c$ is a suitable constant, $h$ is a generic function operating on the skip branch and $f_l(x_{l-1})$ represents the output of the convolutional branch. As we will detail later in this work, this framework generalizes the most commonly employed skip connections.

We assume that we are able, through a proper initialization, to have zero mean and controlled variance (either backward or forward) for each block $f_l$.

In a typical ResNet architecture, $f_l$ is a sequence of two or three convolutions, each one preceded by a ReLU activation - \textit{pre-activation} \citep{he2016identity} - allowing to control both mean and variance through initialization schemes (\ref{eq:he_for}) and (\ref{eq:he_back}).
Note that \textit{post-activated} ResNets do not allow $f_l$ to have zero (either gradient or activation) mean, which corroborates the analysis done by \cite{he2016deep}.

We perform the analysis in this general setting, deriving the condition $h$ and $c$ must satisfy in order to preserve either the forward or backward variance. Then, we propose different ways in which $h$ and $c$ can be defined to satisfy such conditions.

\subsection{The Forward Case}
Let us assume that $\mathbb{E}[x_{0}] = 0$ and $\text{Var}[x_{0}] = 1$, being $x_0$ the input data, and let us reason by induction. 

By the inductive step we assume $\mathbb{E}[x_{l-1}] = 0$ and $\text{Var}[x_{l-1}] = 1$; if weights of each block $f$ are initialized following rule (\ref{eq:he_for}), we can easily verify that $$\mathbb{E}[f_l(x_{l-1})] = \mathbb{E}[x_{l-1}] = 0, \quad \text{Var}[f_l(x_{l-1})] = \text{Var}[x_{l-1}]=1.$$

Recalling \cite{shao2020normalization}, we are allowed to assume that $f_l(x_{l-1})$ and $h(x_{l-1})$ have zero correlation, thus, getting
\begin{align*}
	\mathbb{E}[x_l] & = c\cdot(\mathbb{E}[h(x_{l-1})]) + \mathbb{E}[f_l(x_{l-1})]) \\ 
	& = c\cdot\mathbb{E}[h(x_{l-1})], \\
	\text{Var}[x_l] & = c^2\cdot(\text{Var}[h(x_{l-1})] + \text{Var}[f_l(x_{l-1})]) \\
	& = c^2\cdot (\text{Var}[h(x_{l-1})]+1).
\end{align*}
Thus, defining $h$ so that $\mathbb{E}[h(x_{l-1})] = 0$ and $\text{Var}[h(x_{l-1})] = \frac{1}{c^2}-1$ the activation signal can be preserved and the induction step established.

\subsection{The Backward Case}
Let us assume that for the gradients at the output layer $L$ we have $\mathbb{E}\left[\frac{\partial \mathcal{L}}{\partial x_{L}}\right]=0$ and $\text{Var}\left[\frac{\partial \mathcal{L}}{\partial x_{L}}\right]=C$ and that we initialize  the weight of each block $f_l$ by rule (\ref{eq:he_back}). 

Now, we can assume by induction that the gradients at layer $l$ have zero mean and preserved variance, i.e., $\mathbb{E}\left[\frac{\partial \mathcal{L}}{\partial x_{l}}\right]=0$ and $\text{Var}\left[\frac{\partial \mathcal{L}}{\partial x_{l}}\right]=C$. Since for the gradients at layer $l-1$ we have
\begin{align*}
	\frac{\partial \mathcal{L}}{\partial x_{l-1}} &= \frac{\partial \mathcal{L}}{\partial x_{l}}\frac{\partial x_l}{\partial x_{l-1}}=c\cdot\frac{\partial \mathcal{L}}{\partial x_{l}}\left(\frac{ \partial h(x_{l-1})}{\partial x_{l-1}}+\frac{ \partial f_l(x_{l-1})}{\partial x_{l-1}}\right),
\end{align*}
we get
\begin{align*}
	\mathbb{E}\left[\frac{\partial \mathcal{L}}{\partial x_{l-1}}\right] &=c\cdot \mathbb{E}\left[\frac{\partial \mathcal{L}}{\partial x_{l}}\frac{\partial x_l}{\partial x_{l-1}}\right] \\ &=c\cdot \mathbb{E}\left[\frac{\partial \mathcal{L}}{\partial x_{l}}\right]\mathbb{E}\left[\frac{\partial x_l}{\partial x_{l-1}}\right] = 0.
\end{align*}

Moreover, under the reasonable assumption that there is zero correlation between $\frac{\partial \mathcal{L}}{\partial x_{l}}$ and $\frac{\partial x_l}{\partial x_{l-1}}$, we can further write
\footnotesize
\begin{align*}
	\text{Var}\left[\frac{\partial \mathcal{L}}{\partial x_{l-1}}\right]&=c^2\left(\text{Var}\left[\frac{\partial \mathcal{L}}{\partial x_{l}}\right]\text{Var}\left[\frac{ \partial h(x_{l-1})}{\partial x_{l-1}}+\frac{ \partial f_l(x_{l-1})}{\partial x_{l-1}}\right]\right. \\&\quad+ \text{Var}\left[\frac{\partial \mathcal{L}}{\partial x_{l}}\right]\mathbb{E}\left[\frac{ \partial h(x_{l-1})}{\partial x_{l-1}}+\frac{ \partial f_l(x_{l-1})}{\partial x_{l-1}}\right]^2 \\&\quad\left.+ \mathbb{E}\left[\frac{\partial \mathcal{L}}{\partial x_{l}}\right]^2\text{Var}\left[\frac{ \partial h(x_{l-1})}{\partial x_{l-1}}+\frac{ \partial f_l(x_{l-1})}{\partial x_{l-1}}\right] \right)\\
	&=c^2\left(\text{Var}\left[\frac{\partial \mathcal{L}}{\partial x_{l}}\right]\left(\text{Var}\left[\frac{ \partial h(x_{l-1})}{\partial x_{l-1}}\right]+\text{Var}\left[\frac{ \partial f_l(x_{l-1})}{\partial x_{l-1}}\right]\right)\right. \\&\quad+ \text{Var}\left[\frac{\partial \mathcal{L}}{\partial x_{l}}\right]\left(\mathbb{E}\left[\frac{ \partial h(x_{l-1})}{\partial x_{l-1}}\right]+\mathbb{E}\left[\frac{ \partial f_l(x_{l-1})}{\partial x_{l-1}}\right]\right)^2 \\&\quad\left.+ \mathbb{E}\left[\frac{\partial \mathcal{L}}{\partial x_{l}}\right]^2\left(\text{Var}\left[\frac{ \partial h(x_{l-1})}{\partial x_{l-1}}\right]+\text{Var}\left[\frac{ \partial f_l(x_{l-1})}{\partial x_{l-1}}\right]\right) \right).
\end{align*}

\normalsize
Thanks to the initialization rule (\ref{eq:he_back}), it holds $\mathbb{E}\left[\frac{ \partial f_l(x_{l-1})}{\partial x_{l-1}}\right] = 0$ and $\text{Var}\left[\frac{ \partial f_l(x_{l-1})}{\partial x_{l-1}}\right] = 1$. Therefore we can conclude
\begin{equation} \label{eq:fine_conti_back}
	\begin{aligned}
		\text{Var}\left[\frac{\partial \mathcal{L}}{\partial x_{l-1}}\right]=\;&c^2\cdot\text{Var}\left[\frac{\partial \mathcal{L}}{\partial x_{l}}\right]\left(\text{Var}\left[\frac{ \partial h(x_{l-1})}{\partial x_{l-1}}\right]+1\right)\\&+ c^2\cdot\text{Var}\left[\frac{\partial \mathcal{L}}{\partial x_{l}}\right]\mathbb{E}\left[\frac{ \partial h(x_{l-1})}{\partial x_{l-1}}\right]^2.
	\end{aligned}
\end{equation}

The induction step can therefore be established and the preservation of the gradients signal obtained by suitably defined $h$ and $c$. 

We argue that some of the techniques proposed by \cite{brock2021characterizing,brock2021highperformance} to train deep Residual Networks (weight normalization layers, adaptive gradient clipping, etc.) become necessary because initialization (\ref{eq:brock_for}) focuses on the preservation of the forward activation signal while disregarding the backward one. 

Indeed, the correction factor $\gamma_g^2=2/(1-\frac{1}{\pi})$ in \eqref{eq:brock_for} breaks the conservation property of the gradients signal, as opposed to \eqref{eq:he_for}. As we back-propagate through the model, the factor $\gamma_g^2$ amplifies the gradients signal at each layer, so that the gradients at the last layers are orders of magnitude larger than those at the first layers (going from output to input layers), i.e., the network is in a regime of \textit{exploding gradient}. In the section devoted to the numerical experiments we will show the forward and backward behaviour of these nets.

\subsection{Gradients signal preserving setups} \label{sec:proposed_initialization}
It is well know that \textit{exploding gradients} make training hard (from an optimization perspective). Indeed, without further algorithmic or architectural tricks we are unable to train very deep networks. It is important to note that in the seminal analyses from \cite{glorot2010understanding} and \cite{he2015delving} the derivation implied that preserving the forward variance entailed preserving also the backward variance too (at least to some reasonable amount). Indeed forward and backward variance can be equally preserved if, as already noted, for each layer, the number of input and output channels is equal. On the contrary, in the derivation of \cite{brock2021characterizing, brock2021highperformance}, this relationship between forward and backward variance is lost so that conserving the forward variance implies \textit{exploding gradients}.

For this reason, in the following we mainly focus on the backwards signal, which we argue being a more important thing to look at when forward and backward variance are not tightly related. For this reason, we propose three different possible schemes for choosing $c$ and $h$ in \eqref{eq:gen_scheme}. In particular:
\begin{enumerate}
	\item \textbf{scaled identity shortcut (IdShort):} $h(x) = x$, $c=\sqrt{0.5}$.
	
	This choice, substituting in \eqref{eq:fine_conti_back}, leads to
	\begin{align*}
		\text{Var}\left[\frac{\partial \mathcal{L}}{\partial x_{l-1}}\right]&=\frac{1}{2}\cdot\left(\text{Var}\left[\frac{\partial \mathcal{L}}{\partial x_{l}}\right]\cdot 1 + \text{Var}\left[\frac{\partial \mathcal{L}}{\partial x_{l}}\right]\cdot 1\right)\\&=\text{Var}\left[\frac{\partial \mathcal{L}}{\partial x_{l}}\right], 
	\end{align*}
	i.e., the variance of gradients is preserved. As for the activations, we get $\mathbb{E}[x_l]=0$ and $$\text{Var}[x_l]=0.5\cdot(1+1)=1,$$ i.e., activations signal preservation, for all layers where input and output have the same size.
	
	Note that the latter scheme is significantly different from approaches, like those from \cite{zhang2019, soham2020batch, bachlechner2020rezero}, that propose to add a (learnable) scalar that multiplies the skip branch. In fact, in the proposed scheme the (constant) scalar multiplies both branches and aims at controlling the total variance, without biasing the network towards the identity like in the other approaches.
	
	This is the simplest variance preserving modification of the original scheme that can be devised, only adding a constant scalar scaling at the residual block.
	
	\item \textbf{scaled identity shortcut with a learnable scalar (LearnScalar):} $h(x) = \alpha x$, $\alpha$ initialized at $1$, $c=\sqrt{0.5}$.
	In \eqref{eq:fine_conti_back} we again get at initialization
	\begin{align*}
		\text{Var}\left[\frac{\partial \mathcal{L}}{\partial x_{l-1}}\right]&=\frac{1}{2}\cdot\left(\text{Var}\left[\frac{\partial \mathcal{L}}{\partial x_{l}}\right]\cdot 1 + \text{Var}\left[\frac{\partial \mathcal{L}}{\partial x_{l}}\right]\cdot\alpha^2\right)\\&=\text{Var}\left[\frac{\partial \mathcal{L}}{\partial x_{l}}\right],
	\end{align*}
	and similarly as above we also obtain the forward preservation at all layers with $N=\hat{N}$.
	
	\item \textbf{scaled identity shortcut with a $\boldsymbol{(1\times 1)}$-strided convolution (ConvShort):} $h(x) = W_sx$ initialized by (\ref{eq:he_back}), $c=\sqrt{0.5}$. Since we use He initialization on the convolutional shortcut \citep{he2016identity}, we have $\mathbb{E}\left[\frac{\partial h(x_{l-1})}{\partial x_{l-1}}\right] = 0$ and $\text{Var}\left[\frac{\partial h(x_{l-1})}{\partial x_{l-1}}\right] = 1$, hence we obtain in \eqref{eq:fine_conti_back}
	\begin{align*}
		\text{Var}\left[\frac{\partial \mathcal{L}}{\partial x_{l-1}}\right]&=\frac{1}{2}\cdot\left(\text{Var}\left[\frac{\partial \mathcal{L}}{\partial x_{l}}\right]\cdot 2 + \text{Var}\left[\frac{\partial \mathcal{L}}{\partial x_{l}}\right]\cdot0\right).
	\end{align*}
	Again, if we consider the layers with equal size for inputs and outputs, we also get $\mathbb{E}[x_l]=0$ and $\text{Var}[x_l]=0.5\cdot(1+1)=1$.
	
	Note that this setting (without the scale factor) is commonly used in most ResNet architectures when $x_{l-1}$ and $f_l(x_{l-1})$ have not the same pixel resolution (for instance because $f$ contains some strided convolution) or the same number of channels.
\end{enumerate}

\section{Experiments} \label{sec:experiments}
We start the investigation by numerically computing forward and backward variances for the different initialization schemes. We employ the recently introduced Signal Propagation Plots \citep{brock2021characterizing} for the forwards variance and a modification that looks at the gradients instead of the activations for the backwards case.

We employ the ResNet-50 and ResNet-101 architectures to extract the plots.

In particular we extract the plots for 

\begin{itemize}
	\item classical ResNet with He initialization, \textit{fan in} mode (\ref{eq:he_for}) and \textit{fan out} mode (\ref{eq:he_back});
	\item same of the preceding with batch normalization;
	\item ResNet with the three proposed residual summation modifications and their proper intialization to preserve the backwards variance\footnote{Note that, as in the standard implementation, in IdShort and LearnScalar we employ ConvShort when $x$ has not the same pixel resolution or number of channels of $f(x)$.};
	\item  same as the preceding but employing the intialization of \cite{brock2021characterizing}.
\end{itemize}

For all the initialization schemes, we perform the empirical standardization to zero mean and desired variance of weights at each layer, after the random sampling.

\begin{figure}[ht!]
	\centering
	\begin{subfigure}[h]{1.\columnwidth}
		\centering
		\includegraphics[width=0.99\columnwidth]{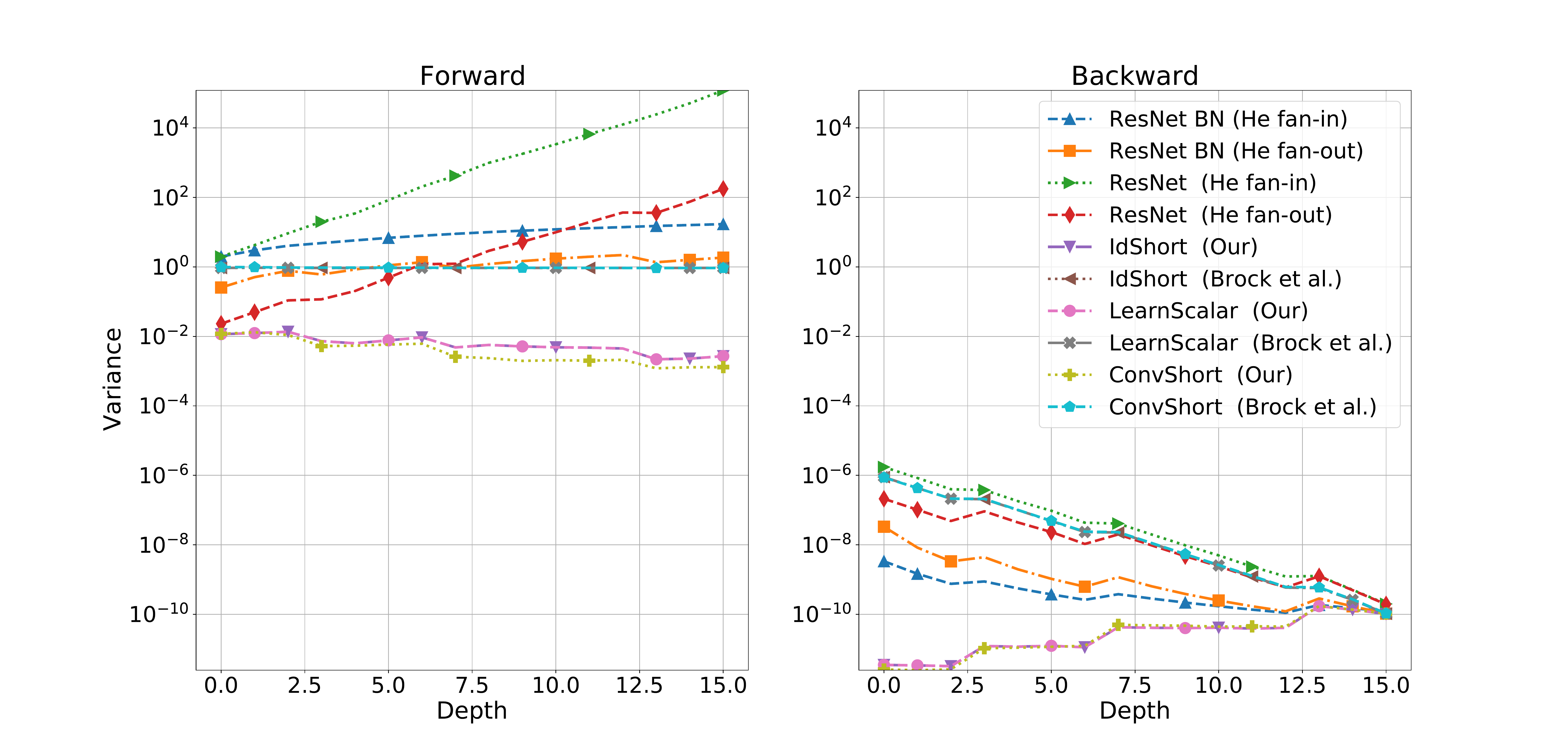}
		\caption{ResNet-50}
		\label{fig:resnet50_spp}
	\end{subfigure} \hfill
	\begin{subfigure}[h]{1.\columnwidth}
		\centering
		\includegraphics[width=0.99\columnwidth]{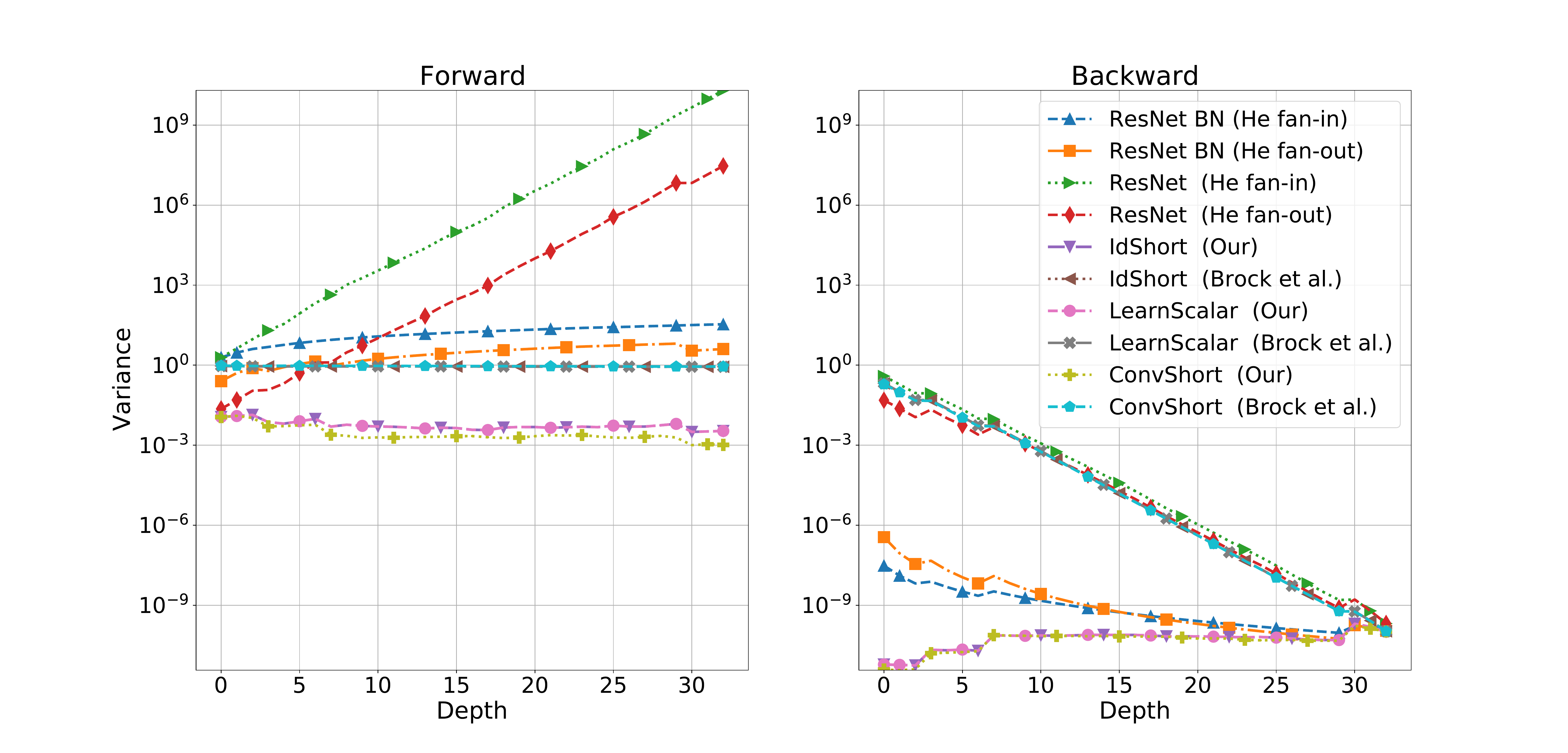}
		\caption{ResNet-101}
		\label{fig:resnet101_spp}
	\end{subfigure} \hfill
	\caption{Signal propagation plots representing the variance of the forward activations (on the left) and the backward gradients variance (on the right) under different initialization schemes: both values refer to residual block output. Values on the $x$-axis denote the residual layer depth, while on the $y$-axis the variance of the signal is reported in a logarithmic scale.}
	\label{fig:resnet_spp}
\end{figure}

From Figure \ref{fig:resnet_spp} we first note that, as already pointed out by \cite{brock2021characterizing}, classical ResNets with He initialization do not preserve neither forwards nor backwards signals while the use of batch normalization manages to fix things up. Interestingly, we note that the observed trends are more conspicuous in deeper networks.

Next, we note that employing the proposed strategies (with proper initialization) we are able to conserve the variance of the gradients. On the contrary, the initialization proposed by \cite{brock2021characterizing} amazingly preserves the forward signal but puts the network in a regime of exploding gradients. Namely, the variance of the gradients exponentially increases going from the deepest to the shallowest residual layers. Additionally, we can also note how the proposed strategies also preserve the activations variance, up to some amount, while when employing the scheme of \cite{brock2021characterizing} the relationship between forward and backward variance is lost.

We continue the analysis by performing a set of experiments on the well-known CIFAR-10 dataset \citep{krizhevsky2009learning} in order to understand if an effective training can be actually carried out under the different schemes and compare them in terms of both train and test accuracy. In particular, we are interested in checking out if the proposed schemes can reach batch normalization test performance.

All the experiments described in what follows have been performed using SGD with an initial learning rate of 0.01, a momentum of 0.9 and a batch size of 128 (100 for ImageNet), in combination with a Cosine Annealing scheduler \citep{loshchilov2016sgdr} that decreases the learning rate after every epoch. Moreover, in addition to the standard data augmentation techniques, we have also employed the recently proposed RandAugment method \citep{cubuk2020rand} and, just for ImageNet, the Label Smoothing technique \citep{zhang2021delving}.

\begin{figure}[ht!]
	\centering
	\begin{subfigure}[h]{1.\columnwidth}
		\centering
		\includegraphics[width=0.99\columnwidth]{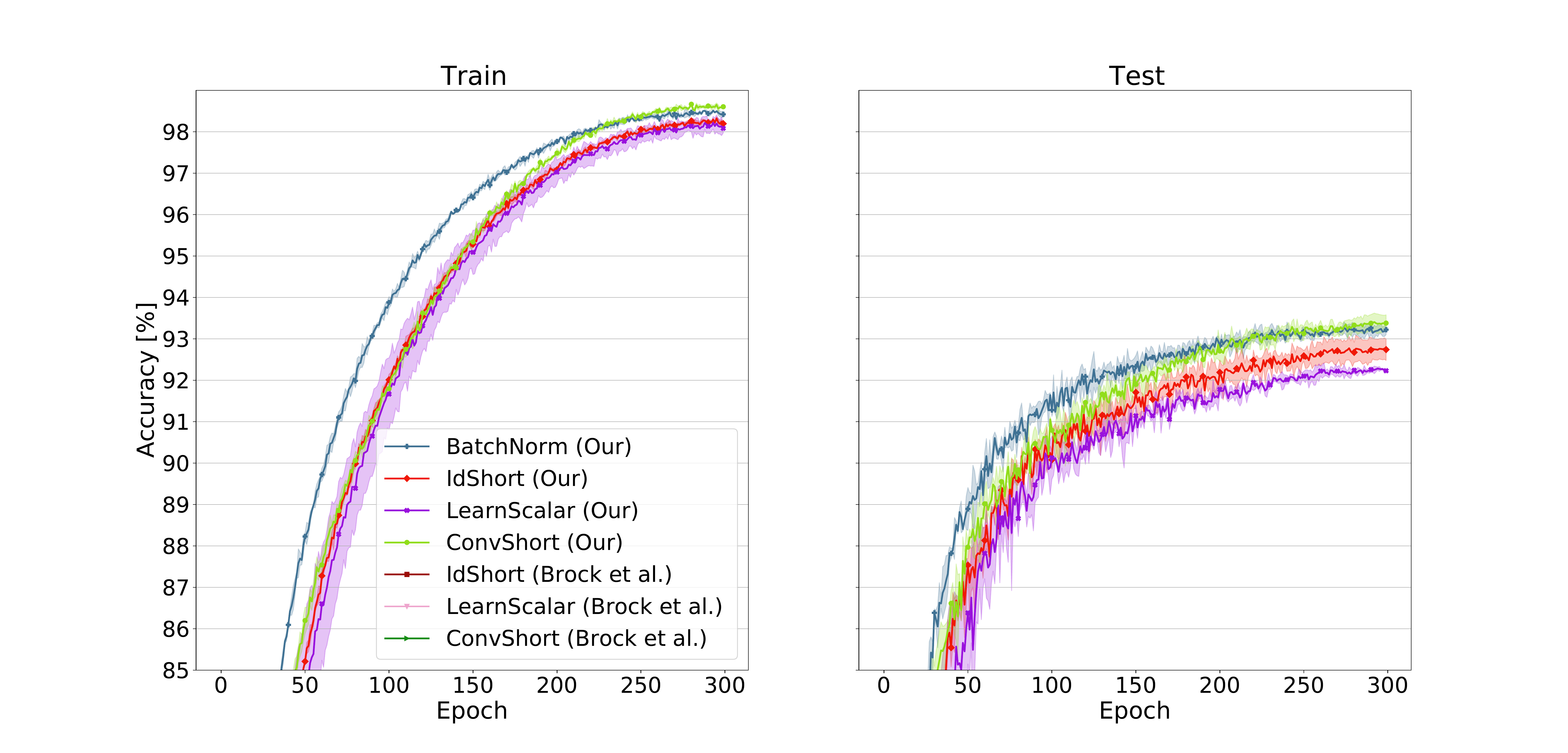}
		\caption{ResNet-50}
		\label{fig:resnet50_accuracies}
	\end{subfigure} \hfill
	\begin{subfigure}[h]{1.\columnwidth}
		\centering
		\includegraphics[width=0.99\columnwidth]{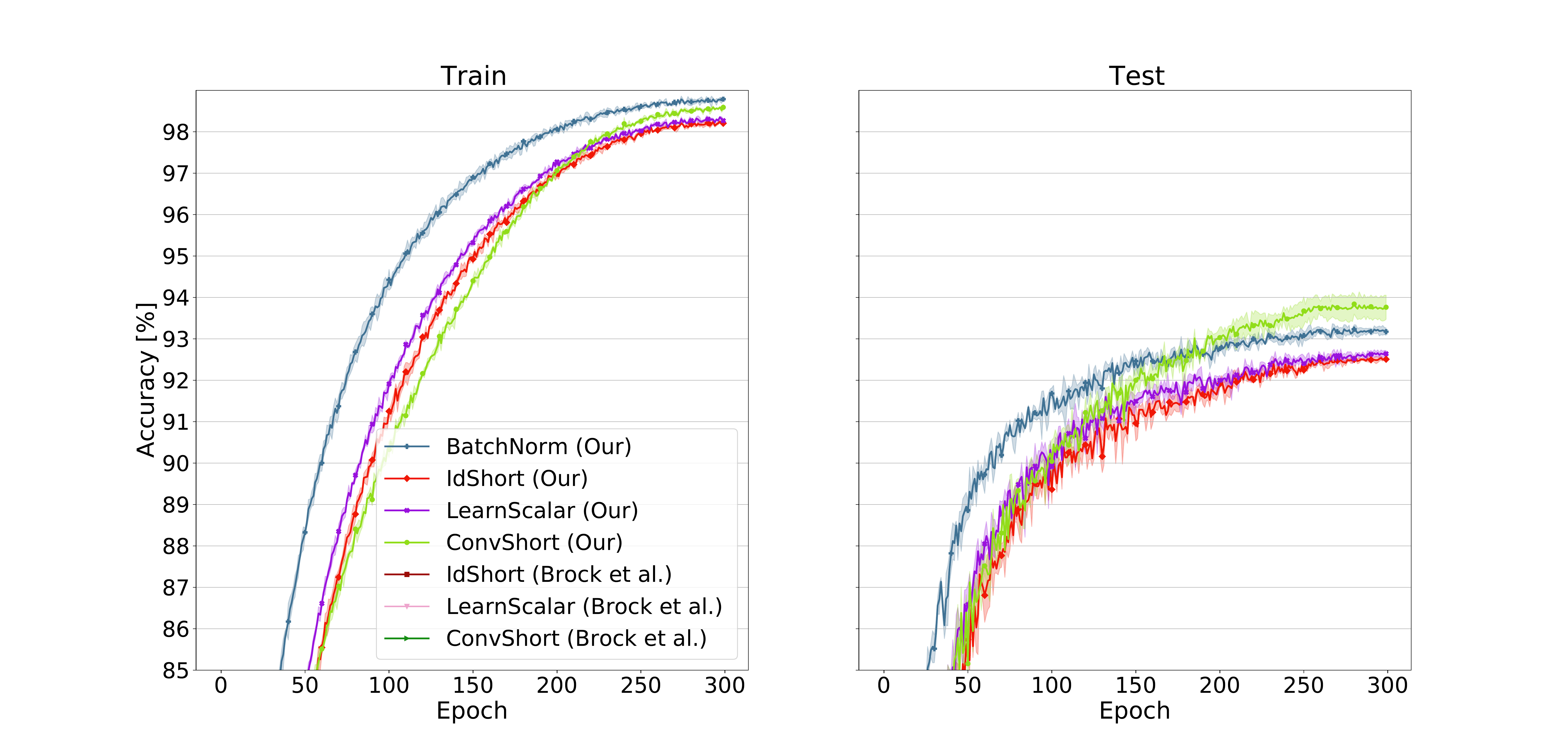}
		\caption{ResNet-101}
		\label{fig:resnet101_accuracies}
	\end{subfigure} \hfill
	\caption{Test and Train accuracies of ResNet on CIFAR-10 dataset under different combinations of residual block modifications and initialization: standard ResNet with BatchNorm and IdShort, LearnScalar, ConvShort using both \cite{brock2021characterizing} and our initialization. Each experiment has been run three times: the solid line is the mean value while the surrounding shadowed area represents the standard deviation. Finally, on the $x$-axis we reprot the epoch at the which the accuracy (in the $y$-axis) has been computed.}
	\label{fig:resnet_accuracies}
\end{figure}

In Figure \ref{fig:resnet_accuracies} both train and test accuracies are shown for all the configurations. The results report the mean and the standard deviation of three independent runs. 

The first thing to notice is that with the initialization scheme of \cite{brock2021characterizing} we are unable to train the network (the curve is actually absent from the plot) for both ResNet-50 and ResNet-101. This is due to the fact that the network, at the start of the training, is in a regime of exploding gradients, as observed in the SPPs. ResNet-18 can be traines using all the considered initialization (see Appendix B).

On the contrary, we can see how, thanks to the correct preservation of the backward signals, training is possible for all the proposed schemes when a gradient preserving initialization scheme is employed.

We also notice that, while all the schemes achieve satisfactory test accuracies, only the \textit{ConvShort} modification has an expressive power able to close  the gap (and even outperform at the last epochs) with the network trained using with Batch Normalization. Thus, according to Figure \ref{fig:resnet50_accuracies} and \ref{fig:resnet101_accuracies}, \textit{ConvShort} appears to be an architectural change that, in combination with the proposed initialization strategy, is able to close the gap with a standard pre-activated ResNet with Batch Normalization (it achieves the second-best in ResNet-18, see Appendix B).

To confirm the effectiveness of the proposed method we also considered more resource-intensive settings, where gradient clipping is expected to be necessary. In particular, we considered the well-known datasets CIFAR-100 \citep{krizhevsky2009learning} and ImageNet \cite{deng2009imagenet}. Based on the results obtained with CIFAR-10, we decided to test the most promising among our architectures, namely, \textit{ConvShort} modification. 

\begin{figure}[ht!]
	\centering
	\includegraphics[width=0.99\columnwidth]{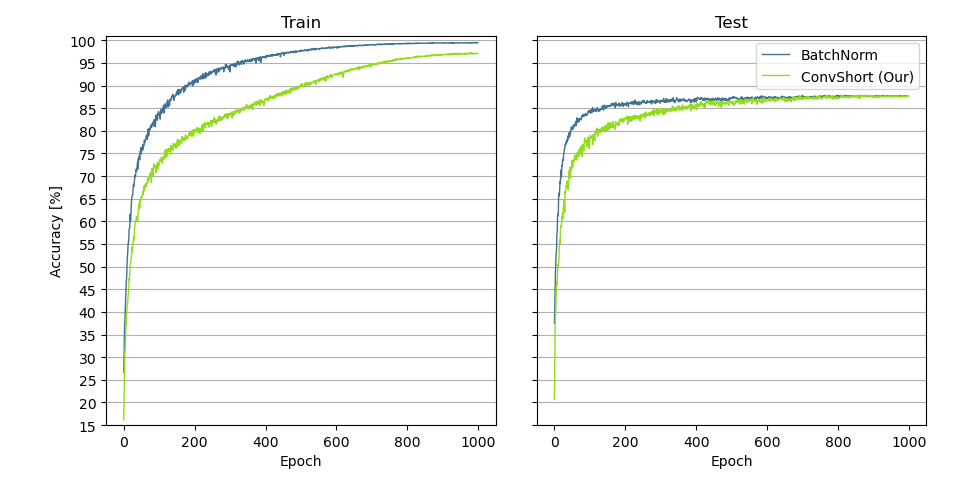}
	\caption{Comparison of Train and Test accuracies of ResNet-50 between standard ResNet with BatchNorm and ConvShort with our initialization using CIFAR-100. Values on $x$-axis denote the epoch at the which the accuracy on the $y$-axis has been computed.}
	\label{fig:rn50_cifar100_accuracies}
\end{figure}

\begin{figure}[ht!]
	\centering
	\includegraphics[width=0.99\columnwidth]{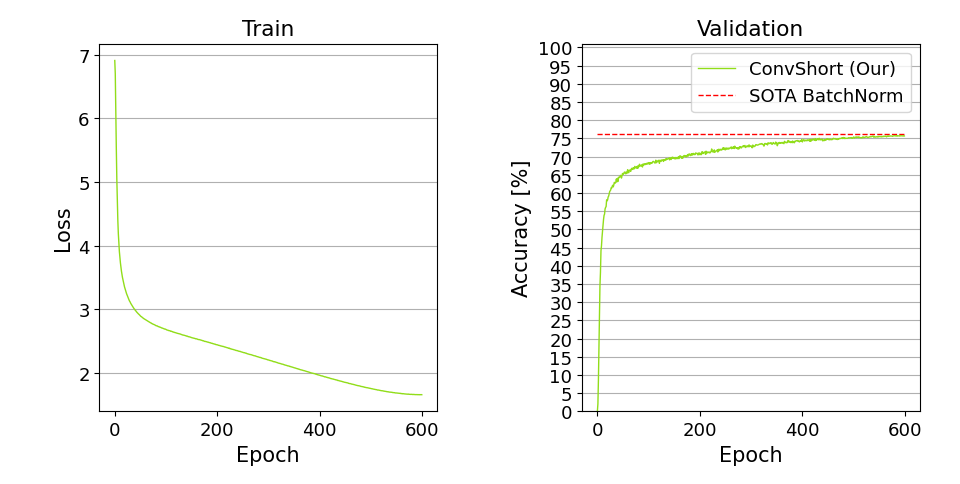}
	\caption{Results obtained training ResNet-50 with our ConvShort modification on ImageNet. Values on $x$-axis denote the epoch at the which the accuracy on the $y$-axis has been computed. The dashed red line is the accuracy reported by PyTorch \cite{NEURIPS2019_9015} for a standard ResNet-50 trained on ImageNet.}
	\label{fig:rn50_imagenet_accuracies}
\end{figure}

In Figure \ref{fig:rn50_cifar100_accuracies} we report the results obtained using our ShortConv modification and a standard ResNet-50 with BatchNormalization. As it is possible to see, training is slower for our setup, but the performance gap eventually closes and testing accuracy of our approach becomes even slightly superior at the end of the process. 
In Figure \ref{fig:rn50_imagenet_accuracies} we show the results obtained with our ShortConv on the well-know ImageNet dataset. In order to evaluate the soundness of our proposal, we compare our results with the accuracy, reported on PyTorch \cite{NEURIPS2019_9015}, reached by a standard ResNet-50 trained on ImageNet. We can observe that the performance obtained with our architecture is in line with the state-of-the-art.

The overall trend seems to indicate that DNNs can be trained up to state-of-the-art performance even without BN, even if this might come at the cost of a slightly longer training; moreover, a strong data augmentation might be needed to compensate the lack of the implicit regularization effects of BN. 

To conclude, we report the number of parameters and FLOPs for the considered architecture in Table \ref{tab:computational_burden_models}. It is important to note that, despite \textit{ConvShort} and \textit{BatchNorm} have the same computational cost, our proposed method have some desirable characteristics (like the independence between the examples in a mini-batch). Moreover, the others configurations can be employed as more light-weight alternatives.

\begin{table}[htb]
	\centering
	\resizebox{\columnwidth}{!}{%
		\begin{tabular}{|lccc|}
			\hline
			\multicolumn{1}{|c}{\textbf{Model}} & \multicolumn{1}{c}{\textbf{Input Resolution}} & \multicolumn{1}{c}{\textbf{Params (M)}} & \multicolumn{1}{c|}{\textbf{\#FLOPs (G)}} \\
			\hline\hline
			ResNet-50 BatchNorm & $32 \times 32 \times 3$ & 38.02 & 4.2 \\
			ResNet-50 IdShort & $32 \times 32 \times 3$ & 23.47 & 2.6 \\
			ResNet-50 LearnScalar & $32 \times 32 \times 3$ & 23.47 & 2.6 \\
			ResNet-50 ConvShort & $32 \times 32 \times 3$ & 38.02 & 4.2 \\
			\hline
			ResNet-101 BatchNorm & $32 \times 32 \times 3$ & 74.78 & 8.92 \\
			ResNet-101 IdShort & $32 \times 32 \times 3$ & 42.41 & 5.02 \\
			ResNet-101 LearnScalar & $32 \times 32 \times 3$ & 42.41 & 5.02 \\
			ResNet-101 ConvShort & $32 \times 32 \times 3$ & 74.78 & 8.92 \\
			\hline
		\end{tabular}
	}
	\caption{Computational cost and number of parameters of the considered architectures.}
	\label{tab:computational_burden_models}
\end{table}

\section{Conclusion}
In this work we proposed a slight architectural modification of ResNet-like architectures that, coupled with a proper weights initialization, can train deep networks without the aid of Batch Normalization. Such initialization scheme is general and can be applied to a wide range of architectures with different building blocks. Importantly, our strategy does not require any additional regularization nor algorithmic modifications, as compared to  other approaches. We show that this setting achieves competitive results on CIFAR-10, CIFAR-100, and ImageNet. The obtained results are in line with the discussed theoretical analysis. 


\subsection{Acknowledgements}
The authors would like to thank Dr.\ Soham De for kindly explaining to us some crucial aspects of his work. We would also like to thank Prof.\ Fabio Schoen for letting us work on this topic and putting at our disposal the resources of GOL and Prof.\ Andrew D. Bagdanov for his precious help in the refinement of this manuscript.

\bibliographystyle{apalike}
\bibliography{references}

\newpage
\appendix

\section{Deriving Standard Initialization Schemes} \label{app:math}

\subsection{He Initialization} \label{app:he}
Consider the response of each layer $l$
\begin{gather*}
	z_l = g(x_{l-1}),\qquad
	x_l = W_l z_l, 
\end{gather*}
where $x$ is a $k^2c$-by-1 vector that represents co-located $k\times k$
pixels in $c$ input channels, $W_l$ is a $d$-by-$n$ matrix where $d$ is the number of filters and $g(\cdot)$ is a nonlinear activation function. In the following, we will consider the classical ReLU, $g(z) = \max(0, z)$.

Formally, let us consider normally distributed input data $x_{l-1} \sim \mathcal{N}(0, \sigma^2)$. It is well known that with this particular activation we get a Rectified Normal Distribution \cite{socci1998rectified,arpit2016normalization} with central moments:
\begin{equation*}
	\mu_g = \mathbb{E}[g(z_l)] =  \frac{\sigma}{\sqrt{2\pi}}, \qquad\sigma_g^2=
	\text{Var}[g(z_l)] =  \frac{\sigma^2}{2} - \frac{\sigma^2}{2\pi}.
\end{equation*}
From the basic properties of variance we also have
\begin{equation} \label{eq:var_formula}
	\mathbb{E}[g(z_l)^{2}] = \mu_g^2 + \sigma_g^2 =  \frac{\sigma^2}{2}.
\end{equation}

Making the assumption that the weights $W$ (we omit for simplicity the dependency on layer $l$) at a network's layer $l$ are i.i.d.\ with zero mean ($\mu_{W}=0$) and that have zero correlation with the input (hence $\text{Var}[Wg(x_{l-1})] = \sigma_{W}^2\sigma^2_g$, putting $Var[W]=\sigma^2_W$),  we obtain for the output elements
that 
\begin{equation*} \label{eq:mean_of_y}
	\mathbb{E}[x_l] = n_{\text{in}}\mu_{g}\mu_W=0
\end{equation*}
and
\begin{equation} \label{eq:variance_of_y}
	\begin{aligned}
		\text{Var}[x_l] & = n_{\text{in}}[\mathbb{E}[W^{2}z_l^{2}] - (\mathbb{E}[Wz_l])^{2}] \\
		& = n_{\text{in}}[\mathbb{E}[W^{2}]\mathbb{E}[z_l^{2}] - (\mathbb{E}[W]\mathbb{E}[z_l])^{2}] \\
		& = n_{\text{in}}[(\mu_{W}^{2} + \sigma_{W}^{2})(\mu_{g}^{2} + \sigma_{g}^{2}) - \mu_{W}^{2}\mu_{g}^{2}] \\
		& = n_{\text{in}}[\sigma_{g}^{2}(\mu_{W}^{2} + \sigma_{W}^{2}) + \sigma_{W}^{2}\mu_{g}^{2}]\\
		& = n_{\text{in}}[\sigma_{W}^2(\mu_g^2+\sigma_g^2)].
	\end{aligned}
\end{equation}
where $n_{\text{in}}=k^2c$ with $k$ the filter dimension and $c$ the number of input channels (\textit{fan in}). Hence if input has unit variance ($\sigma^2=1$) we obtain output unit variance by initializing $W$ in such a way that 
\begin{equation}
	\text{Var}[W] = \frac{2}{n_{\text{in}}}.
\end{equation}

Similarly we can perform  the analysis w.r.t. the gradients signal. 

For back-propagation, we can also write $$\dd{\cal L}{z_l}=\hat{W}\dd{\cal L}{x_l},$$ where $\mathcal{L}$ is the loss function and $\hat{W}$ is a suitable rearrangement of $W$. If weights $W$ are initialized  with zero mean from a symmetric distribution, $\dd{\cal L}{z_l}$ will also have zero mean. 
We can assume $\dd{\cal L}{x_l}$ and $\hat{W}$ to be uncorrelated.

In addition, $$\dd{\mathcal{L}}{x_{l-1}} = \dd{\mathcal{L}}{x_l}g'(x_{l-1});$$ being $g$ the ReLU, $g'(x_{l-1})$ is either 0 or 1 with equal probability, hence, assuming $g'(x_{l-1})$ and $\dd{\mathcal{L}}{x_l}$ uncorrelated, we get 
\begin{gather*}
	\mathbb{E}\left[\dd{\cal L}{x_{l-1}}\right] = \frac{1}{2}\mathbb{E}\left[\dd{\mathcal{L}}{x_l}\right]=\frac{1}{2}\mathbb{E}\left[\hat{W}\right]\mathbb{E}\left[\dd{\cal L}{x_{l-1}}\right]=0,\\ \mathbb{E}\left[\left(\dd{\cal L}{x_{l-l}}\right)^2\right] = \text{Var}\left[\dd{\cal L}{x_{l-l}}\right] = \frac{1}{2}\text{Var}\left[\dd{\cal L}{x_l}\right].
\end{gather*}
Therefore, we can conclude that
\begin{equation*}
	\begin{aligned}
		\text{Var}\left[\dd{\cal L}{x_{l-1}}\right] &= \frac{1}{2}\text{Var}\left[\dd{\mathcal{L}}{x_l}\right] \\
		&=\frac{1}{2}\text{Var}\left[\hat{W}\dd{\cal L}{x_l}\right]\\&=\frac{n_{\text{out}}}{2}\sigma^2_W\text{Var}\left[\dd{\cal L}{x_l}\right].
	\end{aligned}
\end{equation*}
Thus, the initialization
\begin{equation} 
	\text{Var}[W] = \frac{2}{n_{\text{out}}},
\end{equation}
where $n_{\text{out}}=k^2d$ with $k$ the filter dimension and $d$ the number of output channels (\textit{fan out}), allows to preserve the variance of gradients.

\subsection{Brock Initialization}
In contrast with the analysis from Appendix \ref{app:he}, here initial weights are not considered as random variables, but are rather the realization of a random process. Thus, weights mean and variance are empirical values different from those of the generating random process. In order to satisfy the suitable assumptions of the analysis, weights should be actually re-normalized to have empirical zero mean and predefined variance.

Moreover, the channel-wise responses are analyzed. The derivations in \eqref{eq:variance_of_y} should be revised in order to consider expected value and the variance of any single channel $i$ of the output $x_l$ and to take into account constant $\sigma^2_{W_i}$ and $\mu_{W_i}=0$; specifically, we obtain
\begin{equation*} \label{eq:variance_of_y_brock}
	\begin{aligned}
		\text{Var}[{x_{l}}_{i}] & =  \sum_{j=1}^{n_{\text{in}}}\text{Var}[W_{ij}{z_{l}}_j] 
		=  \sum_{j=1}^{n_{\text{in}}}W_{ij}^{2}\text{Var}[{z_l}_j] \\
		&=n_{\text{in}}\left(\sigma_{g}^{2}\cdot\frac{1}{n_{\text{in}}}\sum_{j=1}^{n_{\text{in}}}W_{ij}^{2}\right)\\
		&=N\sigma_{g}^{2}(\mu_{W_i}^{2} + \sigma_{W_i}^{2})\\&= n_{\text{in}}\sigma_g^2\sigma^2_{W_i},
	\end{aligned}
\end{equation*}
so that we retrieve the following initialization rule to preserve an activation signal with unit variance:
\begin{equation}
	\text{Var}[W_i] = \frac{\gamma_g^2}{n_{in}}, 
\end{equation}
where $\gamma_g^2 = \frac{2}{1-\frac{1}{\pi}}$ for the ReLU activation. Note that if mean and variance are preserved channel-wise, then they are also preserved if the whole layer is taken into account.

\section{Experiments on ResNEt-18}
\begin{figure}[htb]
	\centering
	\includegraphics[width=1\columnwidth]{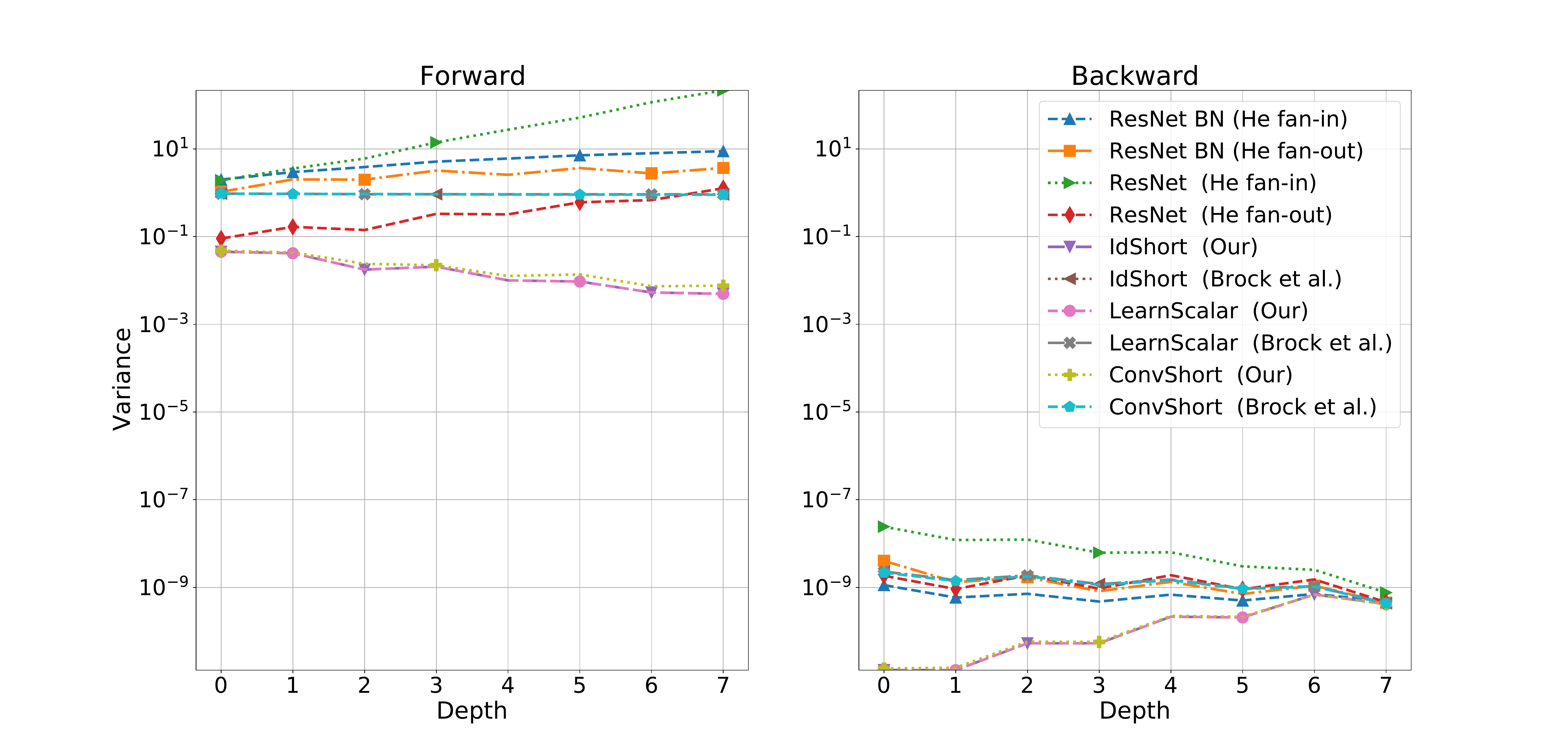}
	\caption{Signal propagation plots representing the variance of the forward activations (on the left) and the backward gradients variance (on the right) under different initialization schemes for ResNet-18: both values refer to residual block output. The $x$-axis is the residual layer depth, while on the $y$-axis the variance of the signal is reported in a logarithmic scale.}
	\label{fig:resnet18_spp}
\end{figure}

\begin{figure}[htb]
	\centering
	\includegraphics[width=1\columnwidth]{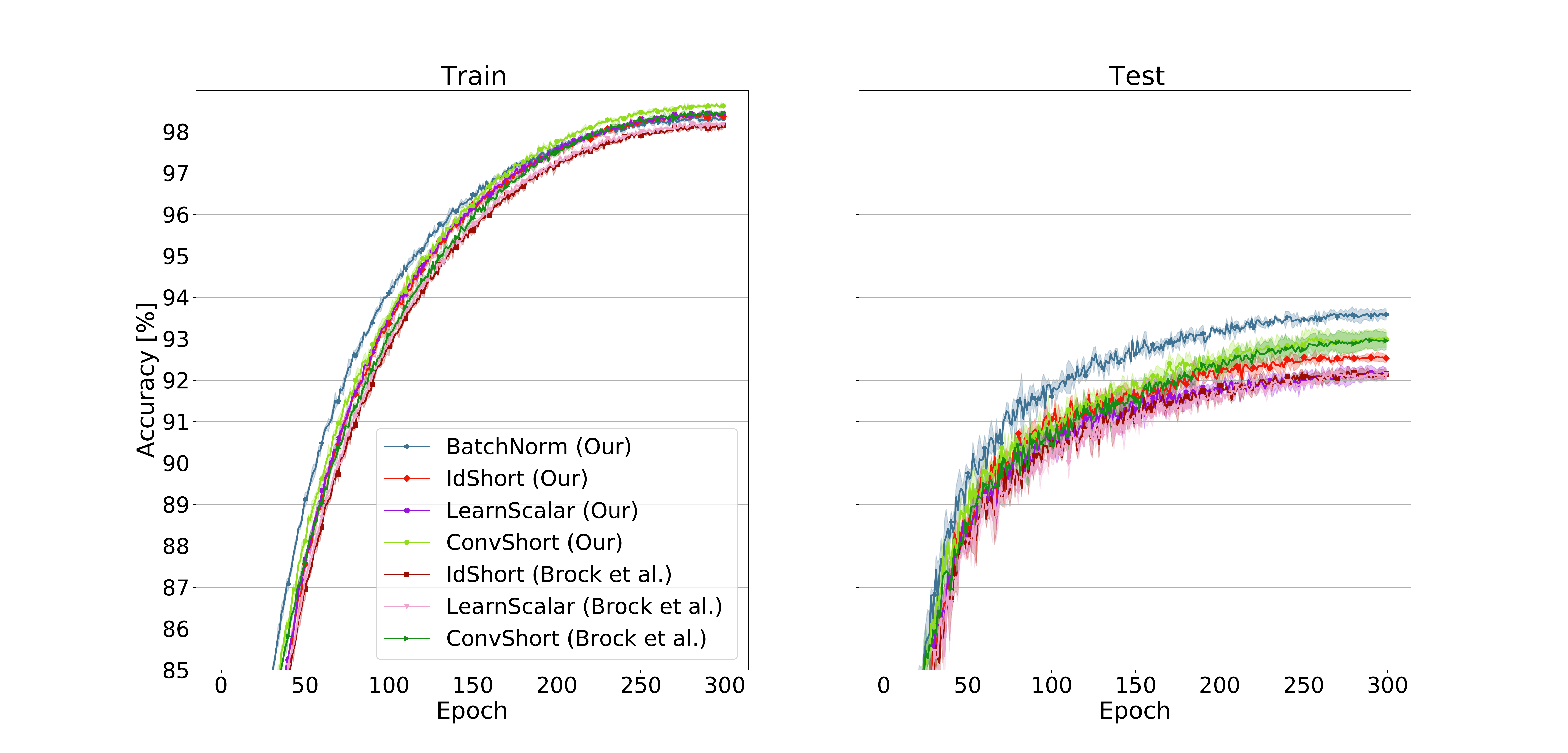}
	\caption{Test and Train accuracies of ResNet-18 under different combinations of residual block modifications and initialization: standard ResNet with BatchNorm and IdShort, LearnScalar, ConvShort using both \cite{brock2021characterizing} and ours initialization. Each experiment has been run three times: the solid line is the mean value while the surrounding shadowed area represents the standard deviation. Finally, the $x$-axis is the epoch at the which the accuracy (reported in the $y$-axis) has been computed.}
	\label{fig:resnet18_accuracies}
\end{figure}

In Fig. \ref{fig:resnet18_spp} and \ref{fig:resnet18_accuracies} we report the results obtained for ResNet-18 with the following configurations

\begin{itemize}
	\item classical ResNet with He initialization, \textit{fan in} mode and \textit{fan out} mode;
	\item same of the preceding with batch normalization;
	\item ResNet with the three proposed residual summation modifications and their proper intialization to preserve the backwards variance\footnote{Note that, as in the standard implementation, in IdShort and LearnScalar we employ ConvShort when $x$ has not the same pixel resolution or number of channels of $f(x)$.};
	\item same as the preceding but employing the intialization of \cite{brock2021characterizing}.
\end{itemize}

Finally, as for the main part of this manuscript, we report the number of parameters and FLOPs for the considered architecture in Table \ref{tab:computational_burden_rn18}. We again highlight that, despite \textit{ConvShort} and \textit{BatchNorm} have the same computational cost, our proposed method have some desirable characteristics (like the independence between the examples in a mini-batch).

\begin{table}[htb]
	\centering
	\resizebox{\columnwidth}{!}{%
		\begin{tabular}{|lccc|}
			\hline
			\multicolumn{1}{|c}{\textbf{Model}} & \multicolumn{1}{c}{\textbf{Input Resolution}} & \multicolumn{1}{c}{\textbf{Params (M)}} & \multicolumn{1}{c|}{\textbf{\#FLOPs (G)}} \\
			\hline\hline
			ResNet-18 BatchNorm & $32 \times 32 \times 3$ & 11.52 & 1.16 \\
			ResNet-18 IdShort & $32 \times 32 \times 3$ & 11.16 & 1.12 \\
			ResNet-18 LearnScalar & $32 \times 32 \times 3$ & 11.16 & 1.12 \\
			ResNet-18 ConvShort & $32 \times 32 \times 3$ & 11.52 & 1.16 \\
			\hline
		\end{tabular}
	}
	\caption{Computational cost and number of parameters of ResNet-18.}
	\label{tab:computational_burden_rn18}
\end{table}

\end{document}